\documentclass{isprs}
\usepackage{setspace}
\usepackage{geometry} 
\usepackage{epstopdf}
\usepackage[labelsep=period]{caption}  

\usepackage{amssymb}
\usepackage{graphicx}
\usepackage{pgfplots}
\usepackage{pgfplotstable}
\usepackage{multirow}
\usepackage{tikz}
\usetikzlibrary{arrows}
\usepackage{amsmath}
\usepackage{makeidx}
\usepackage{subcaption}
\usepackage{color}
\usepackage[normalem]{ulem}
\usepackage{url}
\usepackage{mathtools}
\usepackage{verbatim}
\usepackage{subcaption}

\usepackage{xcolor}
\definecolor{Blue}{RGB}{0,139,249}
\definecolor{Dark green}{RGB}{60,103,28}
\definecolor{Light green}{RGB}{147,169,64}
\definecolor{Grey}{RGB}{160,160,160}
\definecolor{Pink}{RGB}{253,195,252}
\definecolor{Dark orange}{RGB}{255,129,38}
\definecolor{Light orange}{RGB}{255,183,51}
\definecolor{Yellow}{RGB}{255,245,172}

\begin{document}

\title{Combining multiple resolutions into hierarchical representations for kernel-based image classification}

\author{
 Y. Cui\textsuperscript{a}\thanks{International Conference on Geographic Object-Based Image Analysis (GEOBIA 2016), University of Twente in Enschede, The Netherlands.}
 , S. Lef{\`e}vre\textsuperscript{a}, L. Chapel\textsuperscript{a}, A. Puissant\textsuperscript{b}}

\address
{
	\textsuperscript{a }Univ. Bretagne-Sud, UMR 6074, IRISA, F-56000 Vannes, France  \\ \{yanwei.cui, laetitia.chapel, sebastien.lefevre\}@irisa.fr\\
	\textsuperscript{b }Univ. Strasbourg, UMR 7362 LIVE, F-67000 Strasbourg, France   \\ anne.puissant@live-cnrs.unistra.fr\\

}

\icwg{}   

\abstract
{
	Geographic object-based image analysis (GEOBIA) framework has gained increasing interest recently. Following this popular paradigm,  we propose a novel multiscale classification approach operating on a hierarchical image representation built from two  images at different resolutions. They  capture the same scene with different sensors and are naturally fused together through the hierarchical representation, where coarser levels are built from a Low Spatial Resolution (LSR) or Medium Spatial Resolution (MSR) image while finer levels are generated from a High Spatial Resolution (HSR) or Very High Spatial Resolution (VHSR) image. Such a representation allows one to benefit from the context information thanks to the coarser levels, and subregions spatial arrangement information thanks to the finer levels. 
	Two dedicated structured kernels are then used to perform machine learning directly on the constructed hierarchical representation. This strategy overcomes the limits of conventional GEOBIA classification procedures that can handle only one or very few pre-selected scales.
	Experiments run on an urban classification task show that the proposed approach can highly improve the classification accuracy \textit{w.r.t.} conventional approaches working on a single scale. 
}


\maketitle

\section{Introduction}\label{sec:Intro} 

Geographic object-based image analysis (GEOBIA) framework has gained increasing interest recently, especially in the case of very high resolution remote sensing images \cite{Blaschke2014180}. One of the key features for GEOBIA framework is the hierarchical image representation through a tree structure, where objects-of-interest can be revealed through various scales, and where the topological relationship between objects (\textit{e.g.}  A is part of B, or B consists of A) can be easily modeled. In the classification context, however, most papers in literature address one scale only, as being pointed out in a recent survey paper \cite{blaschke2010object}. 

Features extracted from multiple scales are important for improving the object-based classification accuracy, as the underlying tree structure models the hierarchical relationship among the objects \cite{blaschke2010object}. Two important topological information across the scales can be extracted from hierarchical representation: context features and objects spatial arrangement features. 

Context features correspond to the spatial interactions between one region and its surrounding regions. For instance, trees can be classified as residential area instead of forest zone given surrounding regions being buildings and roads.  Such context information can help to disambiguate similar regions during the classification phase \cite{liu2008framework}. Through hierarchical representation, context features can model the evolution of one region and describe it at different levels. Integrating such complementary information leads to some classification accuracy improvement \cite{shackelford2003combined}. Since the spatial position is also implicitly taken into account, it often produces a spatially smoother classification map avoiding ``salt and pepper" effect \cite{bruzzone2006multilevel,lefevre14}. 

Objects spatial arrangement features model the decomposition of an object and the interactions among its subparts. For instance, a residential area is much easier to be identified when knowing it is composed of houses and roads.  Including such information can highly improve the classification rate when spatial interaction between subparts is considered as a critical feature \cite{tang2013multiscale,cui2015subpath}. 

Although features extracted from the multiscale representations are considered as discriminative characteristics for classification, dedicated machine learning algorithms still remain largely unexplored for learning directly from such representations. Recently, advanced machine learning algorithms have been introduced in the GEOBIA framework. Methods such as Support Vector Machine (SVM) \cite{tzotsos2008support}, and Random Forests \cite{stumpf2011object} have been proposed in order to overcome conventional issues of previous GEOBIA classification procedures \cite{shackelford2003combined,benz2004multi}, \textit{e.g.}  manual thresholding and a subjective selection of suitable features. A few dedicated methods have been introduced for taking into account the multiscale features extracted from hierarchical representation \cite{bruzzone2006multilevel}. However, such algorithms able to fully benefit from the multiscale representations remain largely underdeveloped. 

Meanwhile, remote sensing image fusion approaches tend to develop under the GEOBIA framework. These techniques aim to integrate information from different sources, and to produce fused data with more detailed information. For instance, combining high-resolution imagery and LIDAR data allows better accuracy achievements in an urban area classification task \cite{chen2009hierarchical}. As the availability of multi-resolution remote sensing data is rapidly increasing, developing methods able to fuse images from multiple sources and multiple resolutions to improve classification accuracy is becoming  an important topic in remote sensing \cite{zhang2010multi,gomez2015multimodal}. 

\begin{figure*}[htb]
	\centering
	\includegraphics[width=1\textwidth]{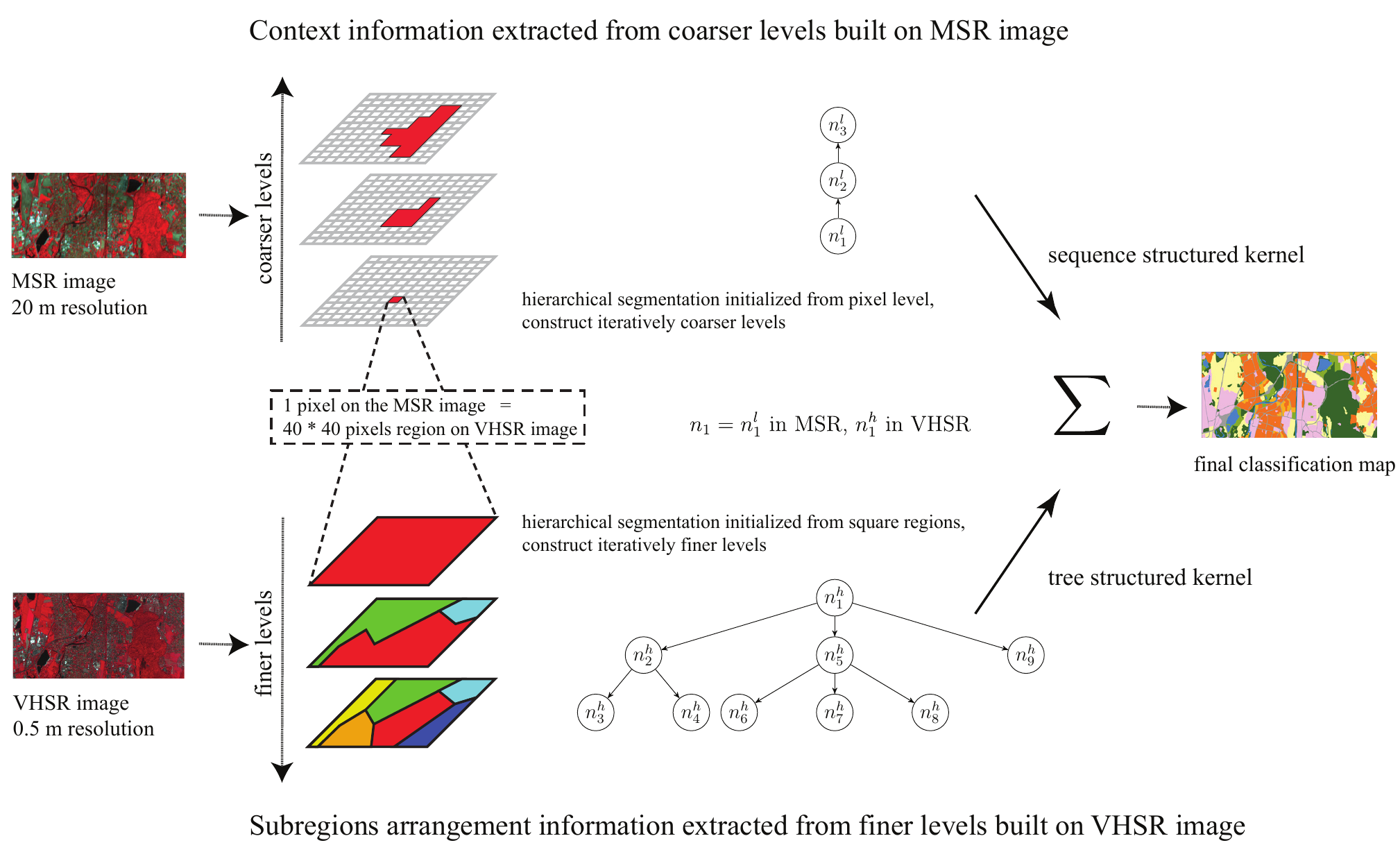}
	\caption{Illustration of the hierarchical image representation for one data instance $n_1$ to be classified. Each data instance  corresponds a pixel of the MSR image $n^l_{1}$, and a $40 \times 40$ square region on the VHSR image $n^h_{1}$. It associates the context information thanks to the coarser levels of the hierarchy built from the MSR image, and the subregion spatial arrangement information thanks to the finer levels constructed on the VHSR image. Both complementary information are taken into consideration thanks to two dedicated structured kernels, then fused together through a composite kernel that provides the classification output.}
	\label{fig:illustration}
\end{figure*}

In this paper, we propose a new approach i) to build a hierarchical image representation from a pair of images with different resolutions (captured with two different sensors) under the GEOBIA framework, and ii) to apply  dedicated kernel methods to perform supervised  classification directly from the constructed tree. 

To build a hierarchical image representation, we rely on two images: a Low Spatial Resolution (LSR) or Medium Spatial Resolution (MSR) image on the one side, and a High Spatial Resolution (HSR) or Very High Spatial Resolution (VHSR) image on the other side. Such a hierarchical representation allows one to benefit of the context information on the coarser levels built from LSR/MSR image, and of the subregions spatial arrangement information on the finer levels built from the HSR/VHSR image. 

To perform image classification from a hierarchical representation, we propose to combine structured kernels computed on two types of structured data: a sequence structured kernel \cite{cui:whispers2016} allows learning the context information with ancestor regions at coarse levels on LSR/MSR image, while a tree structured kernel \cite{cui2015subpath} on HSR/VHSR image makes possible the modeling of the spatial arrangement between subregions. Both kernels exploit complementary information from the hierarchical representation, therefore they are combined at the end. Evaluations show that exploiting  multiscale features through a hierarchical representation with dedicated kernels significantly improves the classification accuracy \textit{w.r.t.} only one single scale.

The paper is organized as follows. We illustrate our main contributions, which include: i) the construction of a hierarchical image representation using two resolution images at different resolutions with different sensors (Sec.~\ref{sec:hiera}), and ii) the kernel to learn directly on the constructed tree (Sec.~\ref{sec:structureLearn}). Then in Sec.~\ref{sec:Evalu}, we detail the experimental setup and discuss the results.  Conclusion and future directions are given at the end of the paper.

\section{Hierarchical representation with multiple resolution images }\label{sec:hiera}

Hierarchical image representation is capable of revealing objects-of-interest through various scales. To construct such representations, one of the most widely adopted techniques is the bottom-up iterative region merging approach \textit{e.g.}  HSeg \cite{tilton2010rhseg}. Starting from the pixel level or any other initial partition (\textit{e.g.} in superpixels), it merges the most similar regions into a new region at each iterative step, until finally the whole image becomes one single region. A threshold parameter  (\textit{e.g.}  a list of similarity criteria following ascending order) is often provided for users to generate the final representation output, with each level being the segmentation map that fulfills the threshold conditions.

Here we build a hierarchical representation with multiple resolution images through two separate steps: i) use LSR/MSR to construct coarser levels of context information on the one side, and ii) use HSR/VHSR image to generate finer levels of subregions spatial arrangement information on the other side, as illustrated in Fig.~\ref{fig:illustration}.

Firstly, we initialize our segmentation at the pixel level on the LSR/ MSR image and construct iteratively the coarser levels. Let $n_{1}$ be a data instance to be classified. Within the  LSR/MSR image, it corresponds to a pixel $n^l_{1}$ and can be represented as a sequence $S= \{n^l_{1},...,n^l_{P}\}$ that models the evolution of the pixel $n^l_{1}$ through the hierarchy. Each node $n^l_{i}$ is described by a $D$-dimensional feature $\boldsymbol{x}^l_{i}$ that encodes the region characteristics, \textit{e.g.} spectral information, size, shape, etc.

Secondly, we use the HSR/VHSR image to provide the fine details of the observed scene for each data instance $n_{1}$. Indeed, one pixel of the LSR/MSR image $n^l_{1}$  always corresponds to a square region of the HSR/VHSR image $n^h_{1}$.  To do so, we initialize the top level of the multiscale segmentation to be the square regions, then construct the finer levels. Through the hierarchy, the data instance $n_{1}$ can be modeled as a tree $T$ rooted in $n^h_{1}$ which encodes subregions and the spatial arrangement among them. The characteristics of region $n^h_{i}$ is also described by a $D$-dimensional feature $\boldsymbol{x}^h_{i}$. 

In the end,  each data instance $n_{1}$ can be represented by an ascending sequence $S$ data from the LSR/MSR image, and a descending tree  $T$ data generated
from the HSR/VHSR image. Learning directly on such representations requires the development of dedicated machine learning algorithms. 

\section{Structured kernels for learning on hierarchical representations}\label{sec:structureLearn}

\subsection{Structured kernels}

To learn from hierarchical representations, we use structured kernels computed on the constructed structures: a sequence structured kernel allows learning context information with ancestor regions at coarser levels on the LSR/MSR image, while a tree structured kernel on the HSR/VHSR image makes possible the modeling of spatial arrangement between subregions. The classification map relies on the composition of both structured kernels. 

Both tree and sequence kernels can be view as instances of the convolution kernel \cite{haussler1999convolution} that defines a general framework to construct structured kernels. It states that a kernel on a complex structure can be formed by tailoring simple kernels computed on its substructures. Formally, let $G,G'$ two structured data and $s,s'$ their substructures, then the kernel between $G,G'$ can be written as: 

\begin{equation}
K(G,G')=\sum_{\substack{s \in G, s' \in G' }} \; K(s,s' ) \;  .
\label{convo}
\end{equation}

\begin{figure}[htb]
\centering
\begin{subfigure}[b]{1\linewidth}
\includegraphics[height = 2cm]{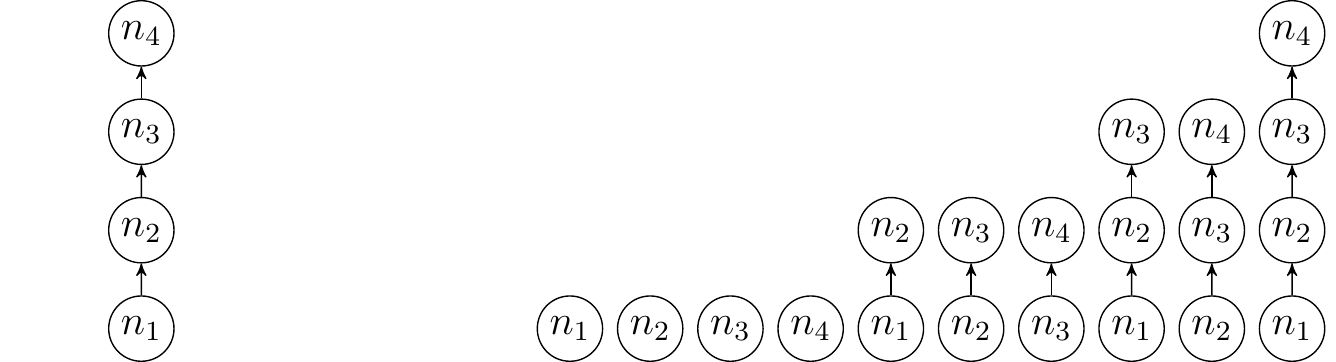}
\caption{A sequence  $S$ and all its subpaths $s$.}
\label{fig:sequence}
\end{subfigure}
\centering
\begin{subfigure}[b]{1\linewidth} 
\includegraphics[height = 1.5cm ]{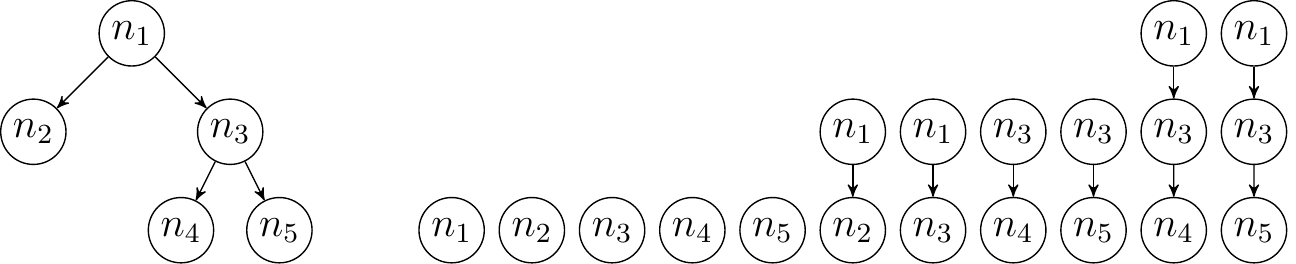}
\caption{A tree  $T$ and all its subpaths $s$.}
\label{fig:subpath}
\end{subfigure}
\caption{Examples of structured data and related substructures.}
\label{fig:struct}
\end{figure}

In order to capture the hierarchical nature of multiscale representation trees and encode the parent-child relationships among the nodes,  subpath substructure has been defined and successfully applied in \cite{cui2015subpath} for tree structured data and in \cite{cui:whispers2016} for sequence structured data. It can be written as $ s = (n_{(1)},n_{(2)}, \cdots n_{(t)}, \cdots n_{(p)})$, $s \in S$, with $(t)$ being the relative position of a node in the subpath, following an ascending order $ 1 \leq t \leq p$, and $p$ being the subpath length. Fig.~\ref{fig:struct} gives an example of a sequence and a tree, with enumeration of all their subpaths $s$ respectively. 

The kernel between two subpaths $s$ and $s'$ with equal length $|s|=|s'|=p$ is defined as the  product of atomic kernels (\textit{e.g.}  Gaussian kernel as in Eq.~\eqref{eq:rbf}) computed on individual nodes $k(n_{(t)},n'_{(t)})$:
\begin{equation} 
K(s,s')= \prod_{\substack{t = 1}}^{p} k(n_{(t)},n'_{(t)} )\;  .
\label{eq:path}
\end{equation}

\subsection{Kernel computation}

We propose here an unified algorithm for computing the sequence and tree kernels based on subpaths. This efficient algorithm can bring down the overall complexity to quadratic \textit{w.r.t.} the size of structures $O(|G||G'|)$. The basic idea is to  iteratively compute  the kernel on subpaths $s$ and $s'$ of length $p$ using previously computed kernels on the subpaths of length $p-1$. The atomic kernel $k(n_{i}, n'_{j})$ between each pair of nodes $(n_i \in G, n'_j \in G')$  thus needs to be computed only once, avoiding redundant computations.

Regarding the sequence kernel, we define a two-dimensional matrix $M$ of size $|S| \times |S'|$, where each element $M_{i,j}$ is computed iteratively as:

\begin{equation} 
M_{i,j} = k(n_i,n'_{j}) (1 + M_{{i-1},{j-1}} ) \;  .
\label{eq:computation}
\end{equation} 
where $M_{i,0} = M_{0,j} = 0$ by convention.

The overall kernel value is then computed as the sum of all the matrix elements.

\begin{equation} 
K(S,S') = \sum_{i=1}^{|S|} \sum_{j=1}^{|S'|} M_{i,j}\;  .
\end{equation}

\begin{figure}[!ht]
	\centering
	\includegraphics[width=.5\textwidth]{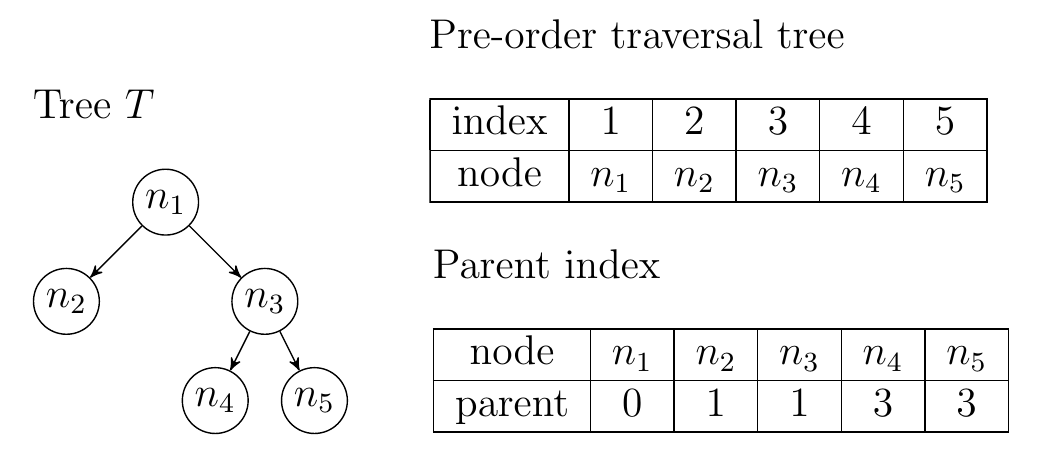}
	\caption{A tree $T$and its associated pre-order depth-first traversal order and parent index table.}
	\label{fig:traversal}
\end{figure}

For the tree kernel, we slightly modify the iteration in Eq.~\eqref{eq:computation} by changing  $ M_{{i-1},{j-1}}$  to $M_{ \text{parent}(n_i), \text{parent}(n'_{j}) }$, where $\text{parent}(n_i)$ refers as parent index of the node $n_i$. It can be constructed by presenting the tree as a sequence of nodes with a pre-order depth-first traversal algorithm \cite{hopcroft1983data}. By convention, the parent index of the root of a tree is $0$, see Fig.~\ref{fig:traversal} for an example.

The overall complexity for both kernels is bounded by the computation of the two-dimensional matrix $M$, which yields $O(|G||G'|)$. 

\begin{figure*}[t]
	\centering
	\begin{subfigure}[b]{0.33\linewidth} 
	\centering		
		\includegraphics[height = 2.2cm ]{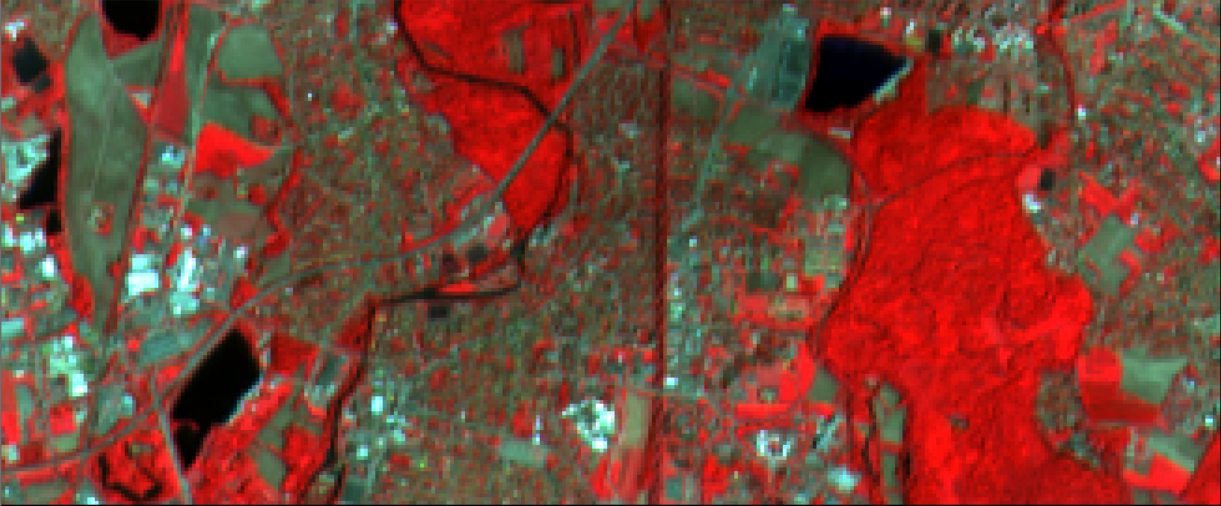}
		\caption{Spot-4 image}
		\label{fig:spot}
	\end{subfigure}
	\centering
	\begin{subfigure}[b]{0.33\linewidth}
	\centering
		\includegraphics[height = 2.2cm]{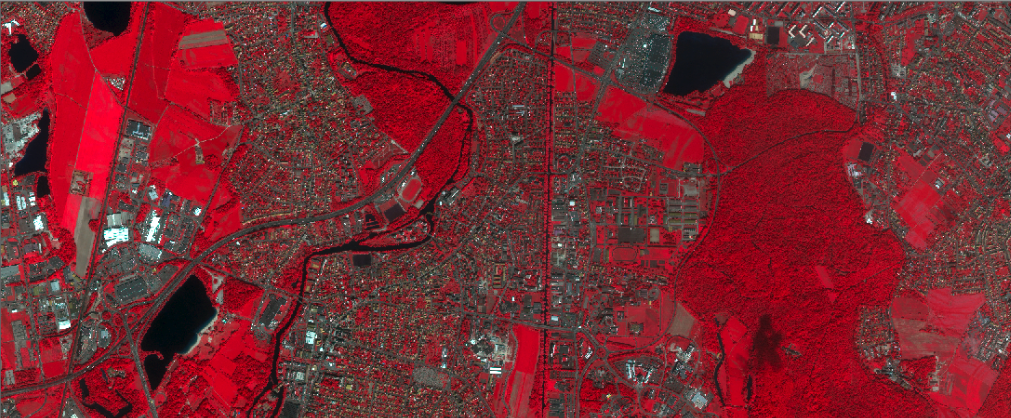}
		\caption{Pleiades image}
		\label{fig:pleiades}
	\end{subfigure}
	\centering
	\begin{subfigure}[b]{0.33\linewidth}
	\centering
		\includegraphics[height = 2.2cm]{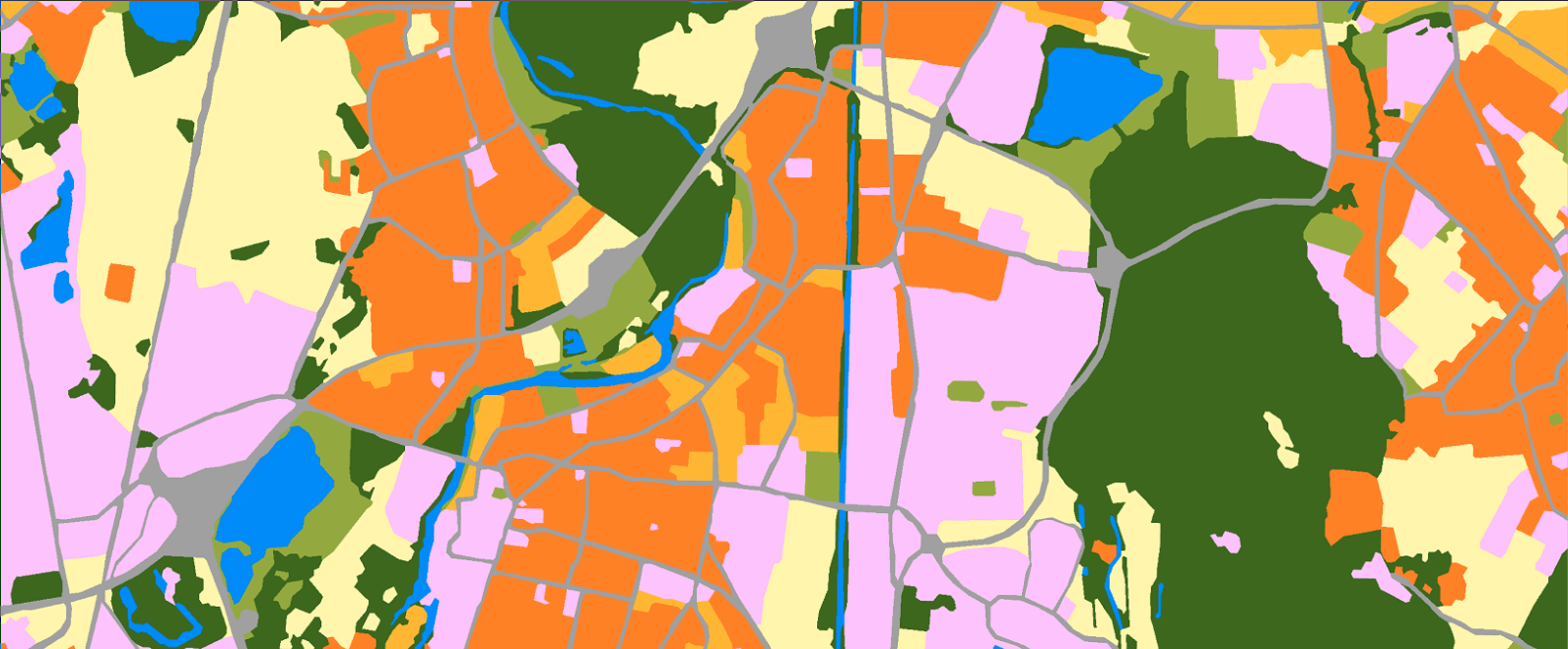}
		\caption{Ground truth image}
		\label{fig:gt}		
	\end{subfigure}
	\caption{Urban scene taken over South of Strasbourg, France. From left to right: false color image of Spot-4 (\textcopyright ~CNES 2012) with 20 m resolution, false color image of Pleiades (\textcopyright ~CNES 2012, distribution Airbus DS / Spot Image) with 50 cm resolution, and the associated ground truth (\textcopyright ~LIVE UMR 7362, adapted from OCSOL CIGAL 2012) with eight thematic classes.}
	\label{fig:rsImage}
\end{figure*} 

\subsection{Kernel combination}

Kernel values must be independent of the size of the structures and should lie in the $(0,1]$ interval.  We thus normalize the kernel value by using the following standard strategy: 

\begin{equation} 
K^*(G,G')= \frac{K(G,G') }{ \sqrt{K(G,G)}\sqrt{K(G',G')}}\;  .
\end{equation}

The final kernel between two data instances $n_1,n'_1$ is computed using a linear combination of the two structured kernels, with a parameter $\rho \in [0,1]$ that controls the importance ratio between the two kernels: 

\begin{equation} 
K(n_1,n'_1) = \rho \times K^*(S,S')   +  (1 - \rho ) \times K^*(T,T') \;  ,
\label{eq:combine}
\end{equation}

where $n_1$ (resp. $n'_1$) is described by $S$ (resp. $S'$) on the LSR/MSR image and $T$ (resp $T'$) on the HSR/VHSR image.

\section{Experiments}\label{sec:Evalu}

\subsection{Study area}

In this paper, we focus on urban land-use classification in the South of Strasbourg city, France. We consider 8 thematic classes of urban patterns as shown in Tab.~\ref{tab:gt} (class details) and in Fig.~\ref{fig:gt} (ground truth image), see \cite{kurtz2012extraction} for more details. Two images from different sources are used:
\begin{itemize}
\item MSR: Spot-4 20 m resolution, 4 bands: Green, Red, NIR, MIR. Image with $326 \times 135$ pixels (Fig.~\ref{fig:spot}).

\item VHSR: Pleiades 0.5 m resolution, 4 bands: Red, Green, Blue, NIR. Image with $13040 \times 5400$ pixels (Fig.~\ref{fig:pleiades}). 
\end{itemize}

\begin{table}[!htb]
	\centering
	\caption{List of classes, their color, and number of pixels in ground truth (on the MSR image, Fig.~\ref{fig:gt}).}
	\begin{tabular}{|l|c|c|}
		\hline
		Class & Color& Nb of pixels\\
		\hline
		Water surfaces&  Blue \textcolor{Blue}{$\blacksquare$ }& 1653   \\
		Forest areas& Dark green \textcolor{Dark green}{$\blacksquare$}& 9315  \\
		Urban vegetation& Light green \textcolor{Light green}{$\blacksquare$} & 1835 \\
		Road & Grey \textcolor{Grey}{$\blacksquare$}& 3498 \\
		Industrial blocks & Pink \textcolor{Pink}{$\blacksquare$}& 8906  \\
		Individual housing blocks & Dark orange \textcolor{Dark orange}{$\blacksquare$} & 9579  \\
		Collective housing blocks & Light orange \textcolor{Light orange}{$\blacksquare$} & 1434  \\
		Agricultural zones& Yellow \textcolor{Yellow}{$\blacksquare$}& 7790  \\
		\hline
		\textbf{Total} &  & \textbf{44010} \\
		\hline
	\end{tabular}
	\label{tab:gt}
\end{table}

\subsection{Experimental setup}

We conduct experiments considering a \textit{one-against-one} SVM classifier, using the Java implementation of LibSVM \cite{chang2011}. The following scenarios are considered:  

\begin{itemize}
\item  Scenario 1: Gaussian kernel at single level on the MSR image \textit{vs.} sequence kernel taking into account the context information at multiple levels on the MSR image. 
\item  Scenario 2: Gaussian kernel at single level on the VHSR image \textit{vs.} tree kernel   taking into account the subregions spatial arrangement information at  multiple levels on the VHSR image. 
\item  Scenario 3: Composite kernel combining both the context and the subregions spatial arrangement information extracted from a hierarchical representation using the two resolution images. 
\end{itemize}

To generate the hierarchical image representation, we rely on HSeg, whose parameters have been empirically fixed as follows:
\begin{itemize}
\item On the MSR image, we generate, from the bottom level of single pixels, 7 additional levels of hierarchical segmentation by increasing the region dissimilarity criteria $\alpha=[2^{-2},2^{-1},...,2^{4}]$. We observe that with such parameters, the number of segmented regions is roughly decreasing by a factor of $2$ between each level. 

\item On the VHSR image, we generate, from the top (root) level of each square region of size $40 \times 40$ pixels (\textit{i.e.} equivalent to a single MSR pixel), 4 additional levels of hierarchical segmentation by decreasing the region dissimilarity criteria $\alpha=[2^{4},2^{3},...,2^{1}]$. Using such parameters, we observe that the number of segmented regions is roughly increasing by a factor of $2$ between each level. 
\end{itemize}

Each region in the hierarchical representation is described by a 8-dimensional feature vector $\boldsymbol{x}$, which includes the region average of the 4 original multi-spectral bands, Soil Brightness index (BI) and NDVI, as well as Haralick texture measurements computed with gray level co-occurrence matrix homogeneity and standard deviation. These features are considered as standard ones in the urban analysis context \cite{forestier2012knowledge}. 

We use Gaussian kernel for the atomic kernel $ k(\cdot,\cdot)$ defined for a pair of nodes $n_i,n'_j$ with respective features $\boldsymbol{x}_i,\boldsymbol{x}'_j$ as 

\begin{equation} 
k(n_i,n'_j) = \exp (-\gamma \lVert \boldsymbol{x}_i-\boldsymbol{x}'_j\rVert^2)\;.
\label{eq:rbf}
\end{equation}

Free parameters are determined by 5-fold cross-validation over potential values: the Gaussian kernel bandwidth $\gamma$ and the SVM regularization parameter $C$.  We also cross-validate the parameter $\rho \in [0,1]$ in Eq.~\eqref{eq:combine} for relative contribution of each kernel. The comparison between different approaches is done by using identical randomly chosen 200 samples per class for training and the rest for testing. All reported results are computed over 10 repetitions of each experiment.

\subsection{Results and discussion}

\begin{table*}[t]
	\centering
	\caption{Classwise accuracies, overall accuracies (OA), average accuracies (AA) and Kappa indices with standard deviation in parentheses. Methods with single level and multiple levels on hierarchical image representation are compared as follows: scenario 1: Gaussian kernel with single level on MSR image (\textbf{single MSR}) \textit{vs.} sequence kernel with multiple levels context information on MSR image (\textbf{context MSR}); scenario 2: Gaussian kernel with single level on VHSR image (\textbf{single VHSR}) \textit{vs.}  tree kernel with  multiple levels subregions spatial arrangement information on VHSR image (\textbf{subregions VHSR}); scenario 3: composite kernel combining both sequence and tree kernel using both MSR and VHSR images (\textbf{composite}). 
	All results are computed over 10 repetitions with best results being boldfaced. Significant differences between single level Gaussian kernels and structured ones using a Wilcoxon test are underlined.}
	\begin{tabular}{|l||c|c||c|c||c|}
		\hline
		Class 						& single MSR& context MSR & single VHSR & subregions VHSR & composite\\
		\hline
		Water surfaces				& 84.90 (2.5) 	& 84.58 (2.2)  & 92.49 (1.3) & 91.69 (1.4)	& 90.40 (1.6)	\\
		Forest areas				& \underline{80.32} (1.5)	& 77.96 (2.1)  & 84.78 (0.8) & \underline{85.80} (0.9)  & 86.76 (1.1)	\\
		Urban vegetation			& 25.99 (4.4) 	& \underline{73.63} (2.1)  & 36.84 (4.9) & 38.16 (3.4)	& 73.19 (2.1)	\\
		Road 						& 38.86 (3.1)	& \underline{43.39} (2.3)  & 48.85 (1.9) & \underline{51.26} (1.7)	& 54.19 (2.3)	\\
		Industrial blocks 			& 35.96 (3.2)	& \underline{70.88} (2.4)  & 23.24 (2.5) & \underline{34.61} (1.9)	& 69.01 (1.6)	\\
		Individual housing blocks 	& 57.09 (4.4) 	& \underline{63.91} (3.3)  & 51.42 (3.3) & \underline{58.02} (2.1)	& 69.62 (1.2)	\\
		Collective housing blocks 	& 24.13 (2.8) 	& \underline{77.89} (3.0)  & 35.32 (3.6) & \underline{38.82} (2.8)	& 79.52 (3.1)	\\
		Agricultural zones			& 36.93 (3.3)	& \underline{67.96} (3.0)  & 67.79 (1.8) & \underline{69.39} (1.7)	& 77.17 (1.9)	\\
		\hline
		\hline
		OA &   51.52 (1.0) & \underline{68.98} (0.9) & 55.91 (0.7)  & \underline{60.53} (0.4) & \textbf{74.47} (0.4) \\
		AA &  48.02 (0.3)& \underline{70.03} (0.5) & 55.09 (0.3) &   \underline{58.47} (0.5)& \textbf{ 74.98} (0.3)\\
		Kappa &  0.426 (0.009)&\underline{ 0.629} (0.009)&   0.485 (0.007) & \underline{0.533} (0.004)&  \textbf{0.693} (0.004)\\
		\hline
	\end{tabular}
	\label{tab:results}
\end{table*}

\begin{figure*}[htb]
	\centering
	\begin{minipage}{0.33\textwidth}
		\begin{subfigure}[b]{1\linewidth} 
			\centering
			\includegraphics[height = 2.2cm]{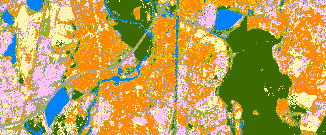}
			\caption{single MSR}
			\label{fig:spot4Pixel}
		\end{subfigure}
		\begin{subfigure}[b]{1\linewidth}
			\centering
			\includegraphics[height = 2.2cm]{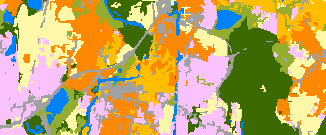}
			\caption{context MSR}
			\label{fig:spot4Context}
		\end{subfigure}
	\end{minipage}
	\begin{minipage}{0.33\textwidth}
		\begin{subfigure}[b]{1\linewidth} 
			\centering
			\includegraphics[height = 2.2cm]{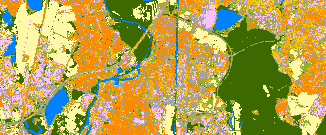}
			\caption{single VHSR}
			\label{fig:PleiadesPixel}
		\end{subfigure}
		\begin{subfigure}[b]{1\linewidth}
			\centering
			\includegraphics[height = 2.2cm]{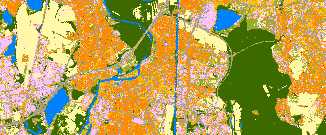}
			\caption{subregions VHSR}
			\label{fig:PleiadesSubregion}
		\end{subfigure}	
	\end{minipage}
	\begin{minipage}{0.33\textwidth}
		\begin{subfigure}[b]{1\linewidth}
			\centering
			\includegraphics[height = 2.2cm]{./image/expe/gt.png}
			\caption{Ground truth image}
			\label{fig:gt2}		
		\end{subfigure}
		\begin{subfigure}[b]{1\linewidth}
			\centering
			\includegraphics[height = 2.2cm]{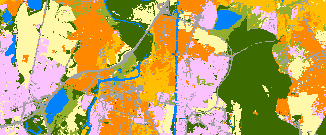}
			\caption{Composite}
			\label{fig:Combined}		
		\end{subfigure}
	\end{minipage}
	\caption{Classification maps for methods using single level and multiple levels of a hierarchical image representation: scenario 1: single level on Spot-4 image \textbf{(a)} \textit{vs.}  multiple levels context information on Spot-4 image \textbf{(b)}; scenario 2: single level on Pleiades image \textbf{(c)} \textit{vs.}  multiple levels subregions spatial arrangement information on Pleiades image \textbf{(d)}; scenario 3: combination of context information  and subregions spatial arrangement information \textbf{(f)}. Ground truth image \textbf{(e)} is also given  as reference.}
	\label{fig:resMap}
\end{figure*}

By taking into account the context information through sequence kernel, the classification results on the MSR image are largely improved comparing to SVM with Gaussian kernel on a single level. We can see in Tab.~\ref{tab:results} that per class accuracy is greatly improved for all classes but two. On the VHSR image, the classification accuracy is improved for all classes but two by using subregions spatial arrangement information. Water surface and urban vegetation classification accuracies remain similar since regions are mostly homogeneous. Moreover, the combination of context information and subregions spatial arrangement information yields an additional improvement, mainly focused on the classes road, individual housing blocks and agricultural zones. 

As shown in Fig.~\ref{fig:spot4Pixel}, the predictions are very noisy with a single level analysis of the MSR image. This is the typical  ``salt and pepper" problem encountered in  remote sensing image classification when the spatial information is not taken into account. Using multiscale information, the spatial dimension is implicitly taken into consideration by the ancestor regions in the hierarchy. Thus a  ``smoother" prediction map can be obtained (as shown in Fig.~\ref{fig:spot4Context}). Let us note that we did not use any post-processing technique to produce such classification map, relying only a structured kernel coping with context information. However, we can also observe that small structures such as road networks disappear in certain areas, and enhance wrongly in other ones. 

As far as the VHSR image is concerned, the prediction maps are noisy with both single and multiple scales. This is due to the fact that the multiscale features extracted on the VHSR image can no longer serve as context information, and spatial relationships among data instances are no longer taken into account. However, it provides the complementary subregions spatial arrangement information, thus leading to a more precise prediction. This conclusion is easier to be reached through quantitative analysis in Tab.~\ref{tab:results}, showing that results improve consistently for most of  the classes (6 out of 8). Classes such as individual housing blocks and industrial blocks are significantly improved, as they can be better characterized by their subregions and the spatial relationships among those regions. Indeed, this shows the advantage of taking into account subregions spatial arrangement information. 

The classification map in Fig.~\ref{fig:Combined} shows that the composite kernel manages to combine the advantages from the two complementary information sources. Indeed, we can observe that the prediction seems to achieve a spatial regularization for the large regions, while providing precision for the small structures such as road networks. Therefore, it leads to the best classification accuracy.

\section{Conclusion}

In this paper, we introduced a novel multiscale approach for combining multiresolution images under the GEOBIA framework. Based on a hierarchical representation generated from images of different resolutions, we propose to use a sequence kernel to take into account the context information built on MSR data, and a tree kernel to capture subregions spatial arrangement information from VHSR data. Both kernels are integrated together through a simple but efficient kernel combination to output final classification results. Evaluations on an urban scene classification problem show that our proposed multiscale approach can significantly improve the classification accuracies \textit{w.r.t.} methods that use only a single spatial scale and only one image.

This paper demonstrates the need of integrating more dedicated machine learning algorithms to take into consideration the topological relationships between objects under the GEOBIA framework. However, the main issue remains the current quadratic kernel computation complexity. In the future, we plan to investigate efficient algorithms, \textit{e.g.}  random Fourier features \cite{bo2009efficient}, to further bring down the computation complexity, and make the proposed approach more adaptable for big remote sensing data.      

\section*{ACKNOWLEDGEMENTS }

The authors acknowledge the support of the French Agence Nationale de la Recherche (ANR) under reference ANR-13-JS02-0005-01 (Asterix project), and the support of R\'egion Bretagne and Conseil G\'en\'eral du Morbihan (ARIA doctoral project).

{
	\begin{spacing}{0.9}
		\bibliography{GEOBIA_YANWEI} 
	\end{spacing}
}

\end{document}